\documentclass[letterpaper, 10 pt, journal, twoside]{IEEEtran}
\IEEEoverridecommandlockouts                              

\usepackage{gensymb}

\usepackage{graphicx} 
\usepackage{amsmath} 

\usepackage{tabularx}
\usepackage{tabularray}

\usepackage[export]{adjustbox}

\usepackage{amsfonts} 

\usepackage{booktabs}
\usepackage{multirow}

\usepackage{hyperref}
\usepackage{svg}

\usepackage{color}
\usepackage{soul}

\begin{document}
©2025 IEEE.  Personal use of this material is permitted.  Permission from IEEE must be obtained for all other uses, in any current or future media, including reprinting/republishing this material for advertising or promotional purposes, creating new collective works, for resale or redistribution to servers or lists, or reuse of any copyrighted component of this work in other works.

The final published version is available on IEEE Xplore: \url{https://ieeexplore.ieee.org/document/11217220}\\
DOI: 10.1109/LRA.2025.3625518

\title{GroundLoc: Efficient Large-Scale Outdoor LiDAR-Only Localization}

\markboth{IEEE Robotics and Automation Letters. Preprint Version. Accepted October, 2025}
{Steinke \MakeLowercase{\textit{et al.}}: GroundLoc: Efficient Large-Scale Outdoor LiDAR-Only Localization} 

\author{Nicolai Steinke$^{1}$ and Daniel Goehring$^{1}$
\thanks{Manuscript received: June 22, 2025; Revised September 15, 2025; Accepted October 6, 2025.}
\thanks{This paper was recommended for publication by Editor Sven Behnke upon evaluation of the Associate Editor and Reviewers' comments.}%
\thanks{This research was funded by the KIS'M project (FKZ: 45AVF3001F), supported by the German Federal Ministry for Digital and Transport (BMDV) under the program: "A future-proof, sustainable mobility system through automated driving and networking."}
\thanks{$^{1}$Dahlem Center for Machine Learning and Robotics (DCMLR), Department of Mathematics and Computer Science, Freie Universit\"{a}t Berlin, Germany, 
{\tt\small \{nicolai.steinke | daniel.goehring\}@fu-berlin.de}}%
\thanks{Digital Object Identifier (DOI): see top of this page.}
}

\maketitle
\begin{abstract}
In this letter, we introduce GroundLoc, a LiDAR-only localization pipeline designed to localize a mobile robot in large-scale outdoor environments using prior maps. GroundLoc employs a Bird's-Eye View (BEV) image projection focusing on the perceived ground area and utilizes the place recognition network R2D2, or alternatively, the non-learning approach Scale-Invariant Feature Transform (SIFT), to identify and select keypoints for BEV image map registration. Our results demonstrate that GroundLoc outperforms state-of-the-art methods on the SemanticKITTI and HeLiPR datasets across various sensors. In the multi-session localization evaluation, GroundLoc reaches an Average Trajectory Error (ATE) well below 50 cm on all Ouster OS2 128 sequences while meeting online runtime requirements. The system supports various sensor models, as evidenced by evaluations conducted with Velodyne HDL-64E, Ouster OS2 128, Aeva Aeries II, and Livox Avia sensors. The prior maps are stored as 2D raster image maps, which can be created from a single drive and require only 4 MB of storage per square kilometer. The source code is available at {\url{https://github.com/dcmlr/groundloc}}.
\end{abstract}
\section{Introduction}
\IEEEPARstart{T}{he} accurate self-localization in large-scale outdoor environments remains an important problem in mobile robotics. As mobile robots, such as autonomous vehicles, operate in open, large-scale areas, there is an increase in both runtime and memory requirements for visual localization systems. Additionally, many assumptions about the environment are not valid in real-world scenarios. This phenomenon is particularly relevant to numerous scenarios encountered by autonomous vehicles in contemporary traffic contexts, such as parking garages, tunnels, and expansive highways. In these situations, satellite-based localization systems, specifically Global Navigation Satellite Systems (GNSS), frequently experience failures and they present significant challenges for various feature-based visual localization methods. These methods typically rely on vertical features for point cloud registration. In repetitive or non-distinctive scenarios lacking stable vertical features, these methods may fail, making them unsuitable for certain environments. While assumption-free methods, such as Iterative Closest Point (ICP)~\cite{besl92}, have been established for some time, the storage requirements for dense 3D point cloud maps remain impractical for mobile robots operating in large-scale outdoor environments. In response to these challenges, we have developed a novel system that leverages learned features or, alternatively, features generated utilizing the Scale-Invariant Feature Transform (SIFT)~\cite{lowe04} from Birds-Eye-View (BEV) images generated from LiDAR data to achieve precise localization of mobile robots in extensive outdoor settings. A ground segmentation algorithm is employed to construct the BEV intensity images, with the ground points serving as the primary data source. This approach enhances the system's resilience to moving and dynamic objects. The prior maps required for this system are compact in size and can be generated from a single visit, thereby simplifying and accelerating the map creation process. Fig.~\ref{fig:motivational_figure} presents the intensity channel of the map generated by the Ouster OS2 128 sensor from the Roundabout HeLiPR~\cite{jung24} sequence, along with the resulting trajectory. This map requires only 5.5~MB of storage while enabling precise LiDAR-only localization. We summarize our main contributions as follows:
\begin{figure}[tpb]
 \centering
 \includegraphics[width=2.4in]{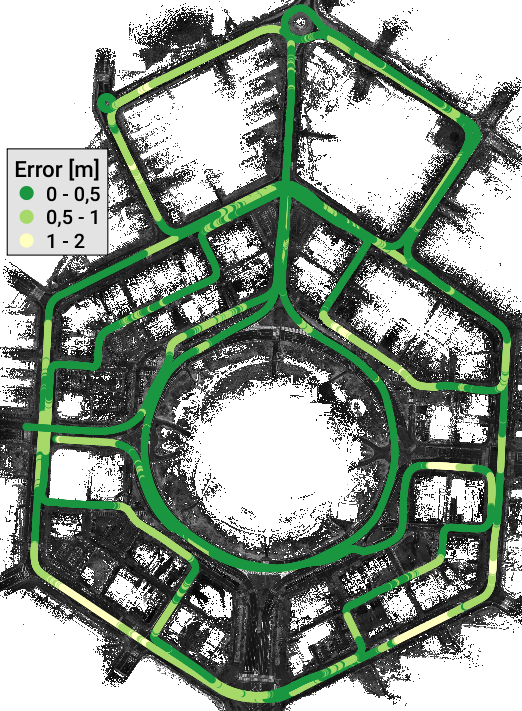}
 \caption{Visualization of multi-session localization results on the HeLiPR Roundabout Ouster sequence of our method. The coloring of the trajectory indicates the translational deviation from the ground truth.}
 \label{fig:motivational_figure}
\end{figure}
\begin{itemize}
    \item We propose an open-source, 3 degrees of freedom (3-DOF) robot localization pipeline for highly precise LiDAR-only localization using prior maps.
    \item We introduce a novel 3-channel BEV image definition and demonstrate the successful application of R2D2 for LiDAR BEV prior map localization.
    \item Our system utilizes 3-channel raster 2D maps that consume only about 4 MB/km$^2$ on average, and it exhibits a high runtime performance exceeding 14 Hz on a consumer laptop using an Ouster OS2 128.
    \item We validate the system in single- and multi-session settings in a wide variety of environments and sensors on the public SemanticKITTI~\cite{behley19} and HeLiPR~\cite{jung24} datasets.
\end{itemize}

\section{Related Work}\label{sec:related_work}
For decades, many autonomous vehicles have relied on LiDAR sensors as part of their perception systems. The active measurement of three-dimensional depth information has established LiDAR sensors as a critical component within sensory pipelines, accelerating their adoption in contemporary autonomous vehicle fleets. Early on, these sensors were utilized not only for perception but also for odometry estimation and localization tasks, often in combination with Global Navigation Satellite System (GNSS) receivers~\cite{levinson10}. The high accuracy, wide field of view (FOV), and three-dimensional data render them a popular choice for mobile robot localization. 
First and foremost, they can be used to determine the robot's own movement, a task called LiDAR odometry estimation, and it represents the first step for most LiDAR localization systems. Notable methods in this field include the LiDAR Odometry and Mapping (LOAM) algorithm~\cite{zhang14} and its variants~\cite{shan18}~\cite{wang21}, which utilize handcrafted features; KISS-ICP~\cite{vizzo23}, which employs the Iterative Closest Point algorithm (ICP)~\cite{besl92} for this purpose; LO-Net~\cite{li19}, and the work by Liu et al.~\cite{liu22} leveraging deep-learning approaches. However, these localization methods are susceptible to drift, which refers to the gradual accumulation of positional and angular errors over time. LiDAR odometry typically serves as the initial step for Simultaneous Localization and Mapping (SLAM) methods, which continuously localize the robot while concurrently constructing a map of the environment. The map registration is often executed using the ICP algorithm or one of its variants~\cite{chen19},~\cite{pan21},~\cite{dellenbach22},~\cite{pan24}. SLAM methods mitigate the drift by employing a technique known as loop closure, where the drift is corrected retroactively when a previously mapped area is revisited during the same run. However, because loop closing can only be applied retroactively, it does not provide a viable solution for online localization of a mobile robot during operation. Alternatively, the drift problem can be addressed through the continuous registration of the sensor measurements against pre-recorded maps, which can be achieved using ICP or other SLAM algorithms.
Nevertheless, the storage of accumulated point cloud maps poses significant challenges due to their substantial size, often consisting of millions or even billions of points.
Given that managing such large volumes of data is impractical for storage-constrained mobile robots, various solutions have been proposed to facilitate more efficient storage of point cloud maps. In addition to compression~\cite{wei19},~\cite{yin21}, and quantization~\cite{chen21},~\cite{yuan22} techniques (such as voxelization, rasterization), the storage of abstract feature maps serves as a strategy to address these storage limitations~\cite{weng18},~\cite{cao20},~\cite{steinke21}. 
Notable works include Weng et al.~\cite{weng18}, who utilized feature maps that focused on pole features for localization. 
Cao et al.~\cite{cao20} enhanced this approach by incorporating additional features, specifically building walls and corners, to improve localization accuracy. Similarly, Steinke et al.~\cite{steinke21} employed the same set of features; however, they derived their maps from publicly available datasets and identified the features through a geometric fingerprinting method. Shi et al.~\cite{shi23} developed a descriptor-based global localization method, Yuan~\cite{yuan22} used voxel maps, and Wolcott and Eustice~\cite{wolcott17} proposed rasterized Gaussian mixture maps to achieve a robust localization system. An alternative approach involves rasterizing the point cloud into a regular range image representation, which can also be used for robot localization purposes, as evidenced by the work of Chen et al.~\cite{chen21}. Many other studies have opted for the bird's-eye view (BEV) quantization method for map creation, capitalizing on its effectiveness in representing spatial information~\cite{levinson10},~\cite{luo22},~\cite{wan18},~\cite{barsan18}.
LiDAR BEV images are generated by rasterizing the point cloud into a regular grid in the xy-plane.  The pixel values in this representation can correspond to various attributes, such as the z-height, intensity value, or other features of the points located within the respective grid cell. 
Many works utilize the intensity or reflectance values of the points as pixel values~\cite{levinson10},~\cite{wan18},~\cite{barsan18}, while others incorporate additional information, including spatial distribution or point count~\cite{wolcott17}, to enhance the representation. In this letter, we will combine compression with BEV quantization to achieve the map storage requirements necessary for mobile robots. We use the intensity value, the slope of the ground, and the z-height variance as the three channels of our BEV images. Regarding the resolution, we demonstrate that precise localization can be achieved with a coarser resolution of 33 cm, in contrast to other state-of-the-art (SOTA) methods that require resolutions of 15 cm~\cite{levinson10} or 5 cm~\cite{barsan18}. The coarser resolution and use of 8-bit values enable our system to utilize significantly smaller maps compared to those employed in other works~\cite{levinson10},~\cite{luo22},~\cite{wan18},~\cite{barsan18}. Many existing methods integrate additional sensors, particularly GNSS and Inertial Measurement Unit (IMU) sensors, with the outputs of LiDAR localization systems~\cite{levinson10},~\cite{cao20},~\cite{steinke21},~\cite{wolcott17},~\cite{wan18},~\cite{barsan18},~\cite{lu19}. Although our approach can be combined with GNSS and IMU sensors to increase accuracy, this work demonstrates the feasibility of a large-scale, LiDAR-only, map-based localization system.
Only a limited number of researchers have explored LiDAR-only systems. Notably, Rozenberszki and Majdik presented LOL~\cite{rozenberszki20}, which combines SegMap~\cite{dube19} with LOAM~\cite{zhang14} to develop a LiDAR-only localization system. Adurthi~\cite{adurthi23} employed a particle filter-based scan matching approach, but could only achieve a frame rate of approximately 2.5 Hz. In contrast, our system achieves a frame rate ranging from 14 Hz to 25 Hz, depending on the sensor used.

\section{GroundLoc Localization Pipeline}\label{sec:terrain_estimation}
\begin{figure}[tpb]
 \centering
 \includegraphics[width=2.2in]{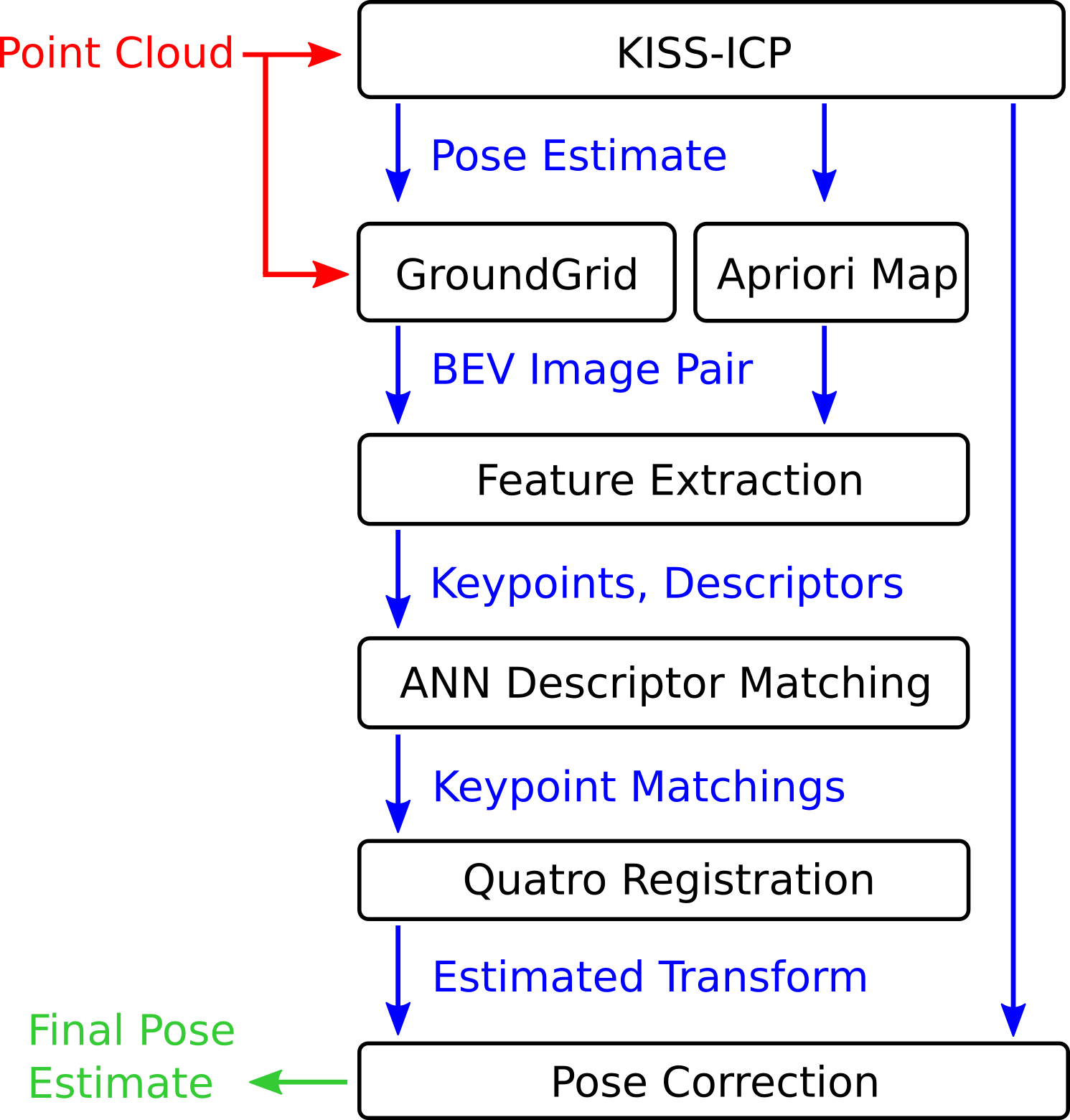}
 \caption{Overview of the proposed system. Input point cloud in red, intermediate results in blue, and output pose estimate in green.}
 \label{fig:overview}
\end{figure}

In this section, we provide a detailed description of our proposed method. Fig.~\ref{fig:overview} illustrates the overall structure of the approach. The input point clouds are segmented and accumulated into a rasterized BEV image using GroundGrid~\cite{steinke24}. Keypoints and descriptors are then extracted using either SIFT~\cite{lowe04} or R2D2~\cite{revaud19} from both the generated BEV image and the reference map. To facilitate efficient matching, we utilize an approximate nearest neighbor KD-Tree implementation to match the source and target descriptors of the keypoints. These matches are further refined by identifying the maximum consensus set using Quatro~\cite{lim22}. Finally, a pose estimate is computed, which is subsequently utilized to correct the odometry estimate.
\subsection{Creation of the LiDAR BEV Images}\label{sec:bev_image_creation}
\begin{figure*}[tpb]
 \centering
 \begin{tabular}{cccc}
     \includegraphics[width=1.63in]{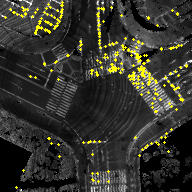} & \includegraphics[width=1.63in]{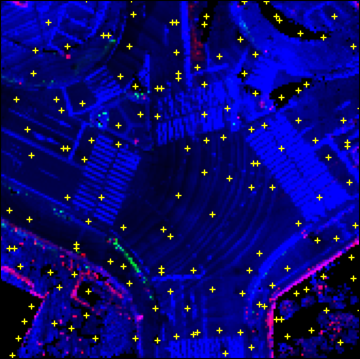} & \includegraphics[width=1.63in]{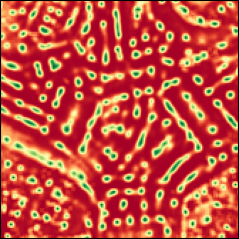} & \includegraphics[width=1.63in]{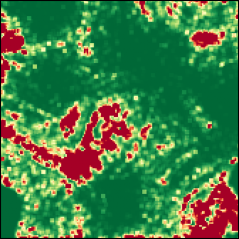}\\
     a) & b) & c) & d)
 \end{tabular}
 \caption{Visualization of the BEV images: a) BEV intensity image with SIFT keypoints; b) 3-channel BEV image with R2D2 keypoints; c) Repeatability map (low-to-high colormap from red to green); d) Reliability map (low-to-high colormap from red to green).}
 \label{fig:bev_image}
\end{figure*}
We employ GroundGrid~\cite{steinke24} to segment the point cloud into ground and non-ground points and to create the BEV images. GroundGrid is a grid mapping method for ground segmentation that allows us to treat the cells of the grid map as pixels in the BEV image. The generated BEV images consist of three channels: intensity, slope, and z-height variance. By exclusively utilizing points classified as ground, we significantly mitigate noise introduced by moving objects. While the intensity channel is commonly utilized for BEV maps, we found it insufficient for robust localization in some scenarios. Therefore, we incorporated a slope channel to represent the local terrain and a height variance channel to preserve information about the roughness of the terrain and static vertical structures. The calculation of the values for each BEV channel is performed as follows:

Let $B^f$ represent our sensor-centered BEV image at frame $f$, which consists of three square matrices of size $s \times s$. These matrices are denoted as $B^f_I$ for intensity, $B^f_S$ for slope, and $B^f_V$ for variance. The elements of the matrices can be accessed by using the indices $(u,v)$, where $u$ and $v$ specify the respective row and column numbers. Similarly, we define $G^f$ as the grid map generated by GroundGrid at frame $f$, which comprises three square matrices of size $s \times s$. These matrices contain the average intensity values, computed ground height, and point z-height variance, and are indexed as $G^f_I$ for intensity, $G^f_S$ for slope, and $G^f_V$ for variance. The elements of these matrices can be accessed by the indices $(u,v)$, identically as the matrices in $B^f$. The grid map is also indexed sensor-centered, but the cell positions are fixed in the world coordinate system and do not move with the robot.
The three matrices, or channels, of the BEV image $B^f$ are related to $G^f$ as follows:
\begin{equation}\label{eq:bev_intensity} 
    B^f_{I}(u,v) = G^f_I(u,v) \cdot I_c
\end{equation}
As shown in~\ref{eq:bev_intensity}, the intensity matrix $B^f_I$ of $B^f$ is equivalent to the intensity matrix $G^f_I$ of GroundGrid, except for the multiplication with the intensity correction factor $I_c$.\\
The slope channel is determined by assessing the terrain height gradient between neighboring cells. Rather than averaging, this channel employs the element-wise minimal slope compared with older observations, further reducing noise from dynamic obstacles:
\begin{equation}~\label{eq:slope_bev}
    B^f_{S}(u,v) = \min \lbrace G^k_S(u,v) | G^k_S(u,v) \neq 0; k = 1,2,...,f \rbrace \cdot S_c
\end{equation} 
The slope channel of a single observation is determined by assessing the relative deviation of the normal vector from absolute vertical, as shown in Equation ~\ref{eq:slope}.
\begin{equation}\label{eq:slope}
    G^f_S(u,v) = 1 - \frac{N_z(u,v)}{|\vec N(u,v)|}
\end{equation}
Here $N(u,v)$ denotes the normal vector at the coordinates $(u,v)$, while $N_z(u,v)$ represents the z-component of this vector. This calculation is conducted exclusively for grid map cells where GroundGrid assigned a ground confidence of $> 0.5$. Otherwise, $G^f_S(u,v)$ is set to zero.\\
\begin{equation}\label{eq:variance_bev}
    B^f_{V}(u,v) = G^f_V \cdot V_c
\end{equation}
Analogous to the intensity channel of the BEV $B^f_I$, the variance channel is calculated by multiplying the grid map's variance values by a normalization factor $V_c$.
To account for the sparsity of the data, the GroundGrid accumulates it as long as it remains within the coverage of the grid $G$. This accumulation is performed as a weighted average based on the count of ground points, with the exception of the slope layer, where the minimum value is used. The weighting reflects the higher amount of information associated with a higher point count. In the case of the slope layer,  weighting is unnecessary, as the ground confidence value ensures that only high-quality observations are utilized. For further details regarding the calculation of the ground confidence, refer to the GroundGrid publication~\cite{steinke24}.

\subsection{Map Creation} 
The prior maps can be generated from a single drive; however, the creation of these maps requires ground truth poses from which the map is constructed. These poses can be generated using a high-quality offline SLAM algorithm operating on the stored point clouds and possible additional sensors of the drive. With these poses, the prior maps can be created by averaging the BEV images at their respective ground truth locations. The intensity channel is weighted by the inverse of the variance to further mitigate the noise introduced by moving objects. To prevent the influence of extremely low variance values -- typically associated with very low point counts -- from dominating the calculations, the inverse variance is constrained to a maximum factor of 1000.
The maps are stored in the GeoTIFF format, utilizing the format's built-in support for the ZSTD compression. The GeoTIFF file format is widely adopted for georeferenced raster data and is compatible with a diverse array of software for reading and editing. Additionally, it offers built-in capabilities for tiling, making it suitable for potentially globally scaled maps.

\subsection{BEV Images Keypoint and Descriptor Extraction}\label{sec:r2d2}
We utilize the reliable and repeatable keypoint extractor R2D2~\cite{revaud19} or, alternatively, the Scale-Invariant Feature Transform (SIFT)~\cite{lowe04}. R2D2 is a learned keypoint detection and description network comprising fewer than 500K parameters, trained on image correspondences. It identifies keypoints based on their repeatability and reliability. Repeatability measures the consistency with which a keypoint can be located at a pixel under varying conditions, while reliability assesses the uniqueness of a keypoint in relation to other nearby keypoints and thus filters repetitive areas. We selected SIFT as a non-learning keypoint and descriptor extraction method for our pipeline due to its robustness against affine transformations, which is essential for the BEV image localization task. Moreover, the SIFT descriptors, like those extracted by R2D2, are 128-dimensional floating-point values that can be compared using the Euclidean distance function, making them interchangeable. Since SIFT operates exclusively on grayscale images, we utilize only the intensity channel of our BEV images.
Images a) and b) of Fig.~\ref{fig:bev_image} present a BEV image and the keypoints (indicated by yellow crosses) extracted by SIFT and R2D2. It is evident that the SIFT keypoints are predominantly located near road markings and pictograms, whereas the keypoints generated by R2D2 appear more evenly distributed throughout the BEV. The visualizations b) and c) show R2D2's repeatability and reliability maps overlaid on the BEV image, with values color-mapped from red to yellow to green, indicating low to high scores, respectively. The repeatability map provides a clear representation of areas identified as significant by R2D2, with the structure of the road, curbs, and lane markings distinctly recognizable. In contrast, the reliability map does not consistently align with the road layout, as most areas are marked with high reliability scores. However, upon closer examination, certain patterns emerge: particularly occluded areas are assigned very low reliability values. Additionally, noisy vegetational areas (located in the lower right corner) and darker areas lacking distinctive features (center left) exhibit lower scores in this metric. In our implementation, we use the product of both scores as the metric for keypoint quality.

\subsection{Training of R2D2}
We trained R2D2 utilizing BEV image pairs with synthetic distortions and further augmentations. Initially, we generate BEV image pair datasets using ground truth poses, ensuring that the BEV images are at least one meter apart to avoid bias towards stationary parts of the datasets. Subsequently, we crop the corresponding ground truth BEV images from the relevant map of the drive. Finally, we train R2D2 with this image pair dataset, incorporating augmentations that simulate the challenges of LiDAR BEV localization, including noise, rotations, and intensity variations. The intensity variations are achieved by adjusting the brightness and contrast of the intensity channel within the BEV image.

\subsection{Calculating the Position Estimate}
As illustrated in Fig.~\ref{fig:overview}, following the feature extraction, the corresponding descriptors are matched using an approximate nearest neighbor algorithm. The matchings are then filtered by positional distance using a dynamic search radius to eliminate implausible correspondences, thereby accelerating the calculation of the maximum consensus set. The search radius is determined based on the offsets from the previous matchings, leveraging the principle that, in sequential data, small offsets in the past are likely to predict small offsets in the future. To calculate the final position estimate, we employ the quasi-SO(3) estimator Quatro~\cite{lim22}, which utilizes a heuristic maximum consensus set method. Quatro demonstrates greater resistance to outliers and produces superior results compared to RANSAC~\cite{fischler1981}, all while maintaining high runtime performance. However, in scenarios involving a very large number of matching keypoints, the performance of Quatro may degrade significantly. To mitigate this issue, we implement a simple hysteresis mechanism to dynamically scale the maximum allowed feature distance and the maximum keypoint number during runtime.
In order to limit the effect of occasional false matchings, we only apply a portion of the calculated pose correction each frame. The amount of the applied correction depends on the number of inliers determined by Quatro and the current speed. A greater number of inliers indicates a higher confidence in the position estimate, while the dependence on the velocity is justified by the observation that drift increases with higher speeds. The formula for the x-dimension is given in Equation~\ref{eq:pos_correction}, the other dimensions (y and yaw) are analogous.
\begin{equation} \label{eq:pos_correction} 
    p_{x} = \max(\min(\gamma o_x, C), -C)
\end{equation}
In this equation, $p_x$ is the final offset adjustment for the $x$-dimension, which is limited by $C$, $o_x$ denotes Quatro's output offset for $x$, and $\gamma$ is the dampening factor. The dampening factor $0 < \gamma \leq 1.0$ was introduced to mitigate positional oscillations that may arise from quantization errors (i. e., when the true position lies between two pixels of the BEV image). The calculation of the maximum and minimum applied values is shown in Equation~\ref{eq:pos_correction_cap}.
\begin{equation} \label{eq:pos_correction_cap}
    C = \frac{|X_{inliers}| \cdot |\vec{v}|}{f_i}
\end{equation}
Here, $\vec{v}$ is the velocity vector measured in $\frac{m}{s}$, $X_{inliers}$ refers to the inlier set, and $f_i$ is the correction factor. 
\section{Experiments and Evaluation}\label{sec:experiment_analysis}

\subsection{Parameters}
GroundGrid requires parameter tuning for each LiDAR sensor to accommodate variations in point quantity and distribution. The used parameters are displayed in Table~\ref{tab:groundgrid_params}.
\begin{table}[b]
  \centering
   \caption{GroundGrid parameters}
\begin{tabular}{l|c|c|c|c|c} ~\label{tab:groundgrid_params}
      & HDL-64E & OS1 64 & OS2 128 & Aeva & Avia \\
      \hline
     $d_{sf}$ & 0.00010 & 0.00015 & 0.00010 & 0.00010 & 0.00100\\
     $t_{minv}$ & 0.0005 & 0.0001 & 0.0005 & 0.0005 & 0.0005\\
     $\theta$ & 5 & 5 & 5 & 5 & 10\\
     $g_{minp}$ & 0.250 & 0.125 & 0.250 & 0.250 & 0.125\\
     $o_{minc}$ & 1.25 & 1.25 & 1.25 & 1.25 & 0.50\\
     $h_g$ & 0.30 & 0.35 & 0.30 & 0.30 & 0.30\\
     $h_o$ & 0.10 & 0.15 & 0.10 & 0.10 & 0.10\\
     $s$ & 20 & 10 & 20 & 20 & 40\\
     $o_t$ & 0.1 & 0.1 & 0.1 & 0.1 & 0.1\\
     $d_{ps}$ & 20.0 & 15.0 & 30.0 & 20.0 & 7.5\\
     $d_{pv}$ & 0.2 & 0.1 & 0.2 & 0.2 & 0.2\\
     $v_{np}$ & 10 & 10 & 10 & 10 & 10\\
     \bottomrule
\end{tabular}
\end{table}
For Quatro~\cite{lim22}, we set parameters $\kappa = 1.39$ and $\overline{c} = 0.5$.
In our evaluations on the SemanticKITTI dataset, R2D2 was trained for 7 epochs using the training sequences from the KITTI360 dataset~\cite{liao23}. For the HeLiPR sequences, we initially trained R2D2 for 15 epochs on the Sejong03 sequence of the MulRan dataset~\cite{kim20}, followed by an additional 3 epochs on the Riverside04 sequence of the HeLiPR dataset, specifically for each sensor. The augmentations employed during training included random rotations in the interval $[-180\degree, 180\degree)$, brightness adjustments of up to 0.5, contrast changes of up to 0.3, and random BEV image cropping with a size parameter of 192 pixels. The learning rate was set to 0.0001, the weight decay to 0.0005, and the batch size to 4.
\begin{table}[h]
  \centering
   \caption{Sensor specific BEV image parameters}
\begin{tabular}{l|c|c|c|c|c} ~\label{tab:bev_params}
      & HDL-64E & OS1 64 & OS2 128 & Aeva & Avia \\
      \hline
     $I_c$ & 2.670 & 1.000 & 0.005 & 0.013 & 0.020\\
     $S_c$ & 0.09 & 0.10 & 0.10 & 0.10 & 0.09\\ 
     $V_c$ & 0.35 & 0.35 & 0.35 & 0.35 & 0.35\\
     \bottomrule
\end{tabular}
\end{table}
The position correction factor $f_i$ (as referenced in Eq.~\ref{eq:pos_correction}) was set to 15 for R2D2 and to 25 for the SIFT version, respectively. The dampening factor $\gamma$ was set to 0.3. Table~\ref{tab:bev_params} shows the BEV image normalization factors. The intensity factors $I_c$ were selected based on the 5th to 95th percentiles of the data and reflect the variations in reflectivity reporting among the respective sensors. 

\subsection{Metrics}
For the evaluation of the matching performance, we utilize the average translation and rotational errors. A registration is deemed successful if the translational error is below 2 m and the rotational error is below 5\degree. In our localization experiments, we focus on the absolute trajectory error (ATE) and the absolute rotational error (ARE). Given that our pipeline only provides a 3-DOF pose estimate, we project all trajectories onto the xy-plane and then apply the Umeyama alignment method. Consequently, the ARE is calculated solely based on the yaw or heading angle.

\subsection{Datasets and State-of-the-Art Baseline Methods}
We evaluate our system using the SemanticKITTI~\cite{behley19} and HeLiPR~\cite{jung24} datasets. 
The SemanticKITTI dataset provides improved ground truth poses compared to the original KITTI~\cite{geiger12} dataset. However, it does not provide multiple drives of the same area, which limits its applicability for evaluating multi-session localization. Due to its popularity, we opted to utilize it in a single-session context, where the map is generated from the same drive as the evaluation run. This approach still offers insights into the localization performance under optimal conditions. In contrast, the HeLiPR dataset provides multiple drives with multiple sensors along nearly identical routes. For the HeLiPR sequences Roundabout and Town, we created the prior maps by combining the first two sequences and used the respective third sequence for evaluation. For the Bridge sequence, we used sequence 04 for map creation and sequence 01 for the evaluation. We selected three sensors for our experiments: the Aeva Aeries II, characterized by its narrow field of view; the Livox Avia, which employs a non-conventional scanning pattern; and the Ouster OS2-128, being a dense 360\degree-sensor. We chose to exclude the Velodyne VLP-16C from our experiments due to the partial obstruction of its field of view by the other sensors and its low resolution, which makes it unsuitable for KISS-ICP. As our method is a 3-DOF localization approach, it is unable to correct errors in pitch, roll, and z-height. Unfortunately, KISS-ICP exhibits significant pitch drift when used with sensors that lack a 360\degree~FOV, such as the Aeva Aeries II and the Livox Avia. This limitation necessitated the projection of KISS-ICP outputs into the xy-plane, zeroing roll and pitch while preserving the length of the estimated movement vector for these sensors:
\begin{equation}\label{eq:2d_projection}
p^f_{xy} = p^{f-1}_{xy} + \Delta p_{xy} \cdot \frac{|\Delta p_{xyz}|}{|\Delta p_{xy}|}
\end{equation}
Eq.~\ref{eq:2d_projection} illustrates the projection of the 3D position $p^f_{xyz}$ at frame $f$ onto the 2D position $p^f_{xy}$, where $\Delta p_{xyz}$ and $\Delta p_{xy}$ denote the differences of subsequent positions generated by KISS-ICP. Although this method introduces minor distortions in sloped terrain to the BEV, our findings indicate that GroundGrid effectively mitigates these distortions, particularly on the highway ramps in the Bridge sequence of the HeLiPR dataset. We select the fingerprinting localization method~\cite{steinke21} as the SOTA baseline method. Given that its feature detector only supports 360\degree~rotating LiDARs, we limit the evaluation of this method to the Ouster OS2-128. Additionally, we report the results of KISS-ICP~\cite{vizzo23} and KISS-SLAM~\cite{guadagnio25} as SOTA LiDAR odometry and SLAM methods, along with Point-to-Plane ICP~\cite{chen91} using pre-generated, voxelized (voxel size 2~m) point cloud maps derived from ground truth poses.
\subsection{Matching Evaluation}
\begin{table}[ht!]
  \centering
   \caption{Single-Session Matching Performance Evaluation}
\begin{tabular}{l|cccc} \label{tab:matching_eval_single}
      Method & Sensor & Trans. err. [m] & Rot. err. [\degree] & Success [\%] \\
      \hline
      SIFT & Aeva & 0.69 & 1.42 & 98.46\\
      R2D2 & Aeva & \textbf{0.60} & \textbf{0.85} & \textbf{99.33}\\
      \midrule
      SIFT & Avia & \textbf{0.25} & 0.32 & 99.70\\ 
      R2D2 & Avia & 0.30 & \textbf{0.28} & \textbf{99.80}\\
      \midrule
      SIFT & Ouster & 0.51 & 0.35 & 99.66\\ 
      R2D2 & Ouster & \textbf{0.43} & \textbf{0.25} & \textbf{99.86}\\  
     \bottomrule
\end{tabular}
\end{table}
To assess the BEV image matching performance of our approach, we construct an evaluation dataset in the same manner as the training datasets. Subsequently, we apply random rotations in the interval $[-180\degree, 180\degree)$ to the query images and randomly translate the x and y positions in the interval $(-20m, 20m)$ from which the map images are cropped. The translation and rotation values computed by GroundLoc are then compared with the ground truth distortion values.
We compare the keypoint and descriptor extraction capabilities of R2D2 and SIFT using various sensors from the KAIST04 sequence of the HeLiPR dataset. Table~\ref{tab:matching_eval_single} presents the results of this evaluation. All sensors achieve high success rates, including the sparser ones with a limited FOV. To assess GroundLoc's multi-session matching capabilities, we conduct the same evaluation as a multi-session localization simulation, with the prior map BEV images created from the KAIST06 sequence.
\begin{table}[tb]
  \centering
   \caption{Multi-Session Matching Performance Evaluation}
\begin{tabular}{l|cccc} \label{tab:matching_eval}
      Method & Sensor & Trans. err. [m] & Rot. err. [\degree] & Success [\%] \\
      \hline
      SIFT & Aeva & 11.25 & 45.50 & 42.78\\
      R2D2 & Aeva & \textbf{4.41} & \textbf{17.77} & \textbf{77.65}\\
      \midrule
      SIFT & Avia & 8.05 & 35.65 & 54.15\\ 
      R2D2 & Avia & \textbf{3.47} & \textbf{16.02} & \textbf{81.57}\\
      \midrule
      SIFT & Ouster & 2.97 & 11.69 & 85.10\\ 
      R2D2 & Ouster & \textbf{1.87} & \textbf{5.61} & \textbf{93.05}\\  
     \bottomrule
\end{tabular}
\end{table}
These sequences were recorded in August and January, introducing seasonal and long-term changes that present additional challenges.
The multi-session performance, as shown in Table~\ref{tab:matching_eval}, declines significantly for all sensors, highlighting the increased difficulty associated with multi-session localization.
R2D2 consistently outperforms SIFT in matching ability in this evaluation, particularly when using sparser sensors. 

\subsection{Single Session Localization}
\begin{table}[t]
  \centering
   \caption{Localization accuracy evaluation on the SemanticKITTI dataset}
   \begin{tabular}{l|cc} 
       Method & {Avg. ATE [m]} & {Avg. ARE [°]}\\
       \midrule
     KISS-ICP & 1.87 $\pm 1.92$ & 0.64 $\pm 0.51$\\
     KISS-SLAM & 1.13 $\pm 0.88$ & 0.50 $\pm 0.36$\\
     Prior Map ICP & 0.18 $\pm 0.09$ & \textbf{0.19} $\pm \textbf{0.10}$\\
     Fingerprint & 0.66 $\pm 0.79$ & 0.47 $\pm 0.37$\\
      Ours (SIFT) & 0.15 $\pm 0.10$ & 0.24 $\pm 0.27$\\
      Ours (R2D2) & \textbf{0.14} $\pm \textbf{0.03}$ & 0.21 $\pm 0.13$\\
     \bottomrule
   \end{tabular}
   \label{tab:kitti_accuracy}
 \end{table}
 In these experiments, all methods demonstrate strong performance, as shown in Table~\ref{tab:kitti_accuracy}. GroundLoc achieves the best ATE results without any outliers. Both feature extraction methods, SIFT and R2D2, yield comparable results in terms of ATE and ARE; however, SIFT exhibits a significantly higher standard deviation, whereas R2D2 produces very consistent results. The Fingerprint localization method shows robust performance across most sequences, although it faces challenges in scenarios with a limited number of pole and corner features. KISS-ICP encounters difficulties with longer sequences due to the accumulation of drift over time. KISS-SLAM's loop-closure mechanism mitigates drift, but it cannot completely eliminate its effects. The ICP registration against a prior point cloud map does not suffer from drift and achieves the best ARE results of all methods; however, its ATE results fall short compared to those of GroundLoc.
\subsection{Multi-Session Localization}
\setlength{\tabcolsep}{4pt}
\begin{table*}[h]
  \centering
   \caption{Multi-Session Localization accuracy evaluation on the HeLiPR dataset (ATE [m] / ARE [°])}
   \begin{tabular}{l|ccc|ccc|ccc}
        &  & Aeva &  &  & Avia &  &  & Ouster & \\
       Method & {Round.} & {Town} & {Bridge} & {Round.} & {Town} & {Bridge} & {Round.} & {Town} & {Bridge}\\
       \midrule
     KISS-ICP & 13.47 / 1.81 & 27.19 / 3.96 & 137.78 / 5.97 & 3.87 / 0.86 & 21.88 / 3.39 & 169.38 / 7.38 & 2.15 / 0.58 & 4.64 / 0.91 & 13.04 / 0.79\\
     KISS-SLAM & 3.38 / 0.87 & 9.43 / 1.22 & 63.97 / 4.93 & 0.82 / 0.54 & 2.70 / 0.80 & 21.48 / 2.05 & 0.52 / 0.50 & 0.68 / 0.52 & 13.28 / 1.28\\
     Prior Map ICP & 230.07 / 58.36 & 310.82 / 85.40 & 1195.36 / 71.17 & \textbf{0.44} / \textbf{0.52} & 314.26 / 87.03 & \textbf{0.48} / \textbf{0.48} & \textbf{0.28} / \textbf{0.23} & \textbf{0.35} / \textbf{0.26} & \textbf{0.30} / \textbf{0.15}\\
     Fingerprint & - / - & - / - & - / - & - / - & - / - & - / - & 0.41 / 0.62 & 0.46 / 2.22 & 1.21 / 1.29\\
     Ours (SIFT) & 2.49 / 0.69 & 2.40 / 1.16 & 1.02 / 1.07 & 0.64 / 0.68 & 1.09 / 0.89 & 60.00 / 2.22 & 0.41 / 0.47 & 0.42 / 0.52
     & 0.45 / 0.43\\
     Ours (R2D2) & \textbf{1.03} / \textbf{0.61} & \textbf{1.17} / \textbf{0.81} & \textbf{0.80} / \textbf{0.87} & 0.57 / 0.66 & \textbf{0.54} / \textbf{0.68} & 0.57 / 1.11 & 0.42 / 0.48 & 0.40 / 0.46 & 0.39 / 0.41\\
     \bottomrule
   \end{tabular}
   \label{tab:helipr_accuracy}
 \end{table*}
 \setlength{\tabcolsep}{6pt}
Table~\ref{tab:helipr_accuracy} presents the results of the multi-session localization experiments. The Roundabout and Town scenarios are characterized by a higher feature density, which aids LiDAR localization. In contrast, the Bridge scenario presents a more challenging environment, as it predominantly involves highway driving and long bridges featuring repetitive structures. Furthermore, the higher speeds (up to 90 km/h) contribute to a stronger drift. This is reflected in the results of KISS-ICP, KISS-SLAM, and the Fingerprint localization method. The prior map ICP method can utilize the higher information density to its advantage, especially with the Ouster sensor, achieving the best overall results when using this sensor.
GroundLoc exhibits robust performance across all scenarios, demonstrating its adaptability to varying conditions.
The Fingerprint localization method yields comparable results in the Roundabout and Town sequences; however, it encounters difficulties in the Bridge sequence.
Furthermore, the angular error is greater with this method compared to the others. We attribute this to the absence of an IMU, which the method can optionally utilize. 
Using the Ouster sensor, many results cluster around an ATE of 0.3-0.4 m. We assume that this pattern is attributable to limitations in the ground truth data regarding its multi-session consistency, indicating that these results are approaching the lower achievable bound.
Regarding the other LiDAR sensors, GroundLoc demonstrates remarkable performance, particularly considering their lower data rates and the increased drift of KISS-ICP. Notably, in the Bridge scenario, GroundLoc is able to reduce the ATE of KISS-ICP by more than a factor of 100, achieving ATEs below 1 meter. KISS-ICP consistently fails to limit the ATE below this value in all sequences due to the cumulative nature of drift without prior map correction. KISS-SLAM effectively reduces drift, especially in the Roundabout and Town sequences; however, it also fails to keep the ATE below 1~m in most cases. The prior map ICP achieves the best scores for the Avia Roundabout and Bridge sequences; however, it struggles with occlusions and fails to maintain map tracking for all Aeva sequences and the Avia Town sequence, resulting in uncontrolled drift and an exceedingly high ATE.
When comparing R2D2 with SIFT, we observe that SIFT can achieve comparable and, in some cases, superior results when using the dense Ouster sensor. However, with the sparser Aeva Aeries II and Livox Avia sensors, GroundLoc's accuracy with SIFT features declines, resulting in performance that falls short of the R2D2 version. This ultimately leads to a loss of map tracking for the SIFT version in the Bridge sequence, causing a significantly higher error.
 \subsection{Ablation Studies}
 \begin{table}[b]
  \centering
   \caption{Multi-Session Localization Results HeLiPR Roundabout sequence across models and sensors (ATE [M])}
\begin{tabular}{l|ccc} ~\label{tab:generalization}
        Training & Aeva & Avia & Ouster \\
      \hline
     Fine-Tuned  & \textbf{1.03} & \textbf{0.57} & \textbf{0.42}\\
     MulRan & 1.16 & 0.62 & \textbf{0.42}\\
     KITTI-360 & 1.12 & 0.79 & 0.43\\ 
     \bottomrule
\end{tabular}
\end{table}
Firstly, we analyze the generalization abilities of our model across sensor models and locations. Table~\ref{tab:generalization} compares the multi-session localization performance of models trained solely on MulRan's Sejong03 sequence or the KITTI-360 dataset with the fine-tuned version on different sensors on the Roundabout sequence of the HeLiPR dataset. The generalization ability across locations appears to be robust, as all models yield good results when using the Ouster OS2 128 sensor, which is similar to those used in the training datasets. Notably, the model trained on data from Germany exhibits only minor performance degradation when applied to a South Korean environment on a similar sensor. However, performance discrepancies increase between sensor models, particularly for the Aeva Aeries II and the Livox Avia. 
This evidence leads us to conclude that, while the models generalize globally across street scenes, they encounter difficulties when transitioning between sensor models, especially when there are significant differences in intensity response and scanning patterns.
 \begin{table}[tb]
  \centering
   \caption{Localization Accuracy w/ Ouster Map\\ (ATE [m] / ARE [°])}
\begin{tabular}{l|cccc} ~\label{tab:crossmap}
     Method & Sensor & Round. & Town & Bridge  \\
      \hline
     SIFT & Aeva & 1.47 / 0.63 & 2.66 / 0.99 & 2.83 / 1.18\\
     R2D2 & Aeva & \textbf{1.04} / \textbf{0.59} & \textbf{1.14} / \textbf{0.83} & \textbf{0.93} / \textbf{1.00}\\
     \midrule
     SIFT & Avia & 0.57 / \textbf{0.57} & 1.37 / 0.88 & 33.24 / 1.84\\
     R2D2 & Avia & \textbf{0.53} / 0.58 & \textbf{0.49} / \textbf{0.63} & \textbf{0.71} / \textbf{1.08}\\
     \bottomrule
\end{tabular}
\end{table}
Table~\ref{tab:crossmap} presents the findings of the inter-LiDAR localization with the Aeva Aeries II and the Livox Avia, utilizing a map created with the Ouster OS2 128. The results indicate the capability to localize using a map created by a different, denser sensor, with some of the results surpassing those reported in Table~\ref{tab:helipr_accuracy}. This suggests an advantage of more complete map information and enables the use of more advanced sensors for map creation, which can then be utilized by less sophisticated platforms.

\subsection{Runtime and Map Storage Analysis}
GroundLoc achieved a processing rate exceeding 14 Hz in all experiments using a laptop equipped with an Intel Core i9-13950HX and an NVIDIA GeForce RTX 4090 Mobile. KISS-ICP, GroundGrid, and the BEV registration are executed in parallel to eliminate waiting times. GroundLoc's maps are stored as 3-channel, 8-bit rasters with a resolution of 33 cm per pixel, utilizing ZSTD compression. The Ouster OS2 128 produces the densest maps, which average to 4.09 MB/km$^2$ in our experiments. This storage requirement is significantly lower than that of the Fingerprint localization method, which is reported to require 33.75 MB/km$^2$ when using public datasets~\cite{steinke21}, and that of the downsampled point cloud maps for the ICP method, which require 15.32 MB/km$^2$ utilizing the PCD format with its built-in compression. The storage of the unprocessed point cloud data would need approximately 55.4 GB/km$^2$.

\section{Conclusion}
In this letter, we have introduced GroundLoc, a BEV image LiDAR-only prior map localization pipeline that enables the localization of a mobile robot in large-scale scenarios, relying solely on LiDAR sensors while meeting online runtime constraints. Keypoint and descriptor extraction can be performed using either a conventional method or by means of a CNN. The storage requirements for the maps are minimal; they can be created from a single drive and edited using standard GeoTIFF tools. The pipeline demonstrates the ability to deliver highly accurate localization estimates across a variety of sensors and scenarios, as evidenced by a comprehensive evaluation conducted on several public datasets. To support the robotics research community in addressing the remaining challenges in large-scale robot localization, the source code is made available as open source. 



\begin{thebibliography}{99}
\bibitem{besl92} P. J. Besl and N. D. McKay, ``A method for registration of 3-D shapes," IEEE Trans. on Pattern Anal. and Mach. Intell., vol. 14, no. 2, pp. 239-256, 1992.
\bibitem{lowe04} D. G. Lowe, ``Distinctive Image Features from Scale-Invariant Keypoints," Int. J. Comput. Vis., vol. 60, pp. 91–110, 2004. 
\bibitem{jung24} M. Jung, W. Yang, D. Lee, H. Gil, G. Kim, and A. Kim, ``HeLiPR: Heterogeneous LiDAR dataset for inter-LiDAR place recognition under spatiotemporal variations," Int. J. Robot. Res., vol. 43, no. 12, pp. 1867-1883, 2024.
\bibitem{behley19} J. Behley, M. Garbade, A. Milioto, J. Quenzel, S. Behnke, C. Stachniss, and J. Gall, ``SemanticKITTI: A Dataset for Semantic Scene Understanding of LiDAR Sequences," Proc. IEEE Int. Conf. Comput. Vis., pp. 9297-9307, 2019.
\bibitem{levinson10} J. Levinson and S. Thrun, ``Robust vehicle localization in urban environments using probabilistic maps," Proc. IEEE Int. Conf. Robot. Autom., pp. 4372-4378, 2010.
\bibitem{zhang14} J. Zhang and S. Sanjiv, ``LOAM: Lidar odometry and mapping in real-time," Robotics: Science and systems, vol. 2, no 9, pp. 1-9, 2014.
\bibitem{shan18} T. Shan and B. Englot, ``LeGO-LOAM: Lightweight and Ground-Optimized Lidar Odometry and Mapping on Variable Terrain," Proc. IEEE Int. Conf. Intell. Robots Syst., pp. 4758-4765, 2018.
\bibitem{wang21} C. Wang, C. Wang, C. -L. Chen, and L. Xie, ``F-LOAM : Fast LiDAR Odometry and Mapping," Proc. IEEE Int. Conf. Intell. Robots Syst., pp. 4390-4396, 2021.
\bibitem{vizzo23} I. Vizzo, T. Guadagnino, B. Mersch, L. Wiesmann, J. Behley, and C. Stachniss, ``KISS-ICP: In Defense of Point-to-Point ICP – Simple, Accurate, and Robust Registration If Done the Right Way,"  IEEE Robot. Autom. Lett., vol. 8, no. 2, pp. 1029-1036, 2023.
\bibitem{li19} Q. Li, S. Chen, C. Wang, X. Li, C. Wen, M. Cheng, and J. Li, ``LO-Net: Deep Real-Time Lidar Odometry," Proc. IEEE Conf. Comput. Vis. Pattern Recognit., pp. 8465-8474, 2019.
\bibitem{liu22} T. Liu, Y. Wang, X. Niu, L. Chang, T. Zhang, and J. Liu, ``LiDAR Odometry by Deep Learning-Based Feature Points with Two-Step Pose Estimation," Remote Sensing, vol. 14, 2022.
\bibitem{chen19} X. Chen, A. Milioto, E. Palazzolo, P. Giguère, J. Behley, and C. Stachniss, ``SuMa++: Efficient LiDAR-based Semantic SLAM," Proc. IEEE Int. Conf. Intell. Robots Syst., pp. 4530-4537, 2019.
\bibitem{pan21} Y. Pan, P Xiao, Y. He, Z. Shao, and Z. Li, ``MULLS: Versatile LiDAR SLAM via Multi-metric Linear Least Square," Proc. IEEE Int. Conf. Robot. Autom., pp. 11633-11640, 2021.
\bibitem{dellenbach22} P. Dellenbach, J. -E. Deschaud, B. Jacquet, and F. Goulette, ``CT-ICP: Real-time Elastic LiDAR Odometry with Loop Closure," Proc. IEEE In. Conf. Robot. Autom., pp. 5580-5586, 2022.
\bibitem{pan24} Y. Pan, X. Zhong, L. Wiesmann, T. Posewsky, J. Behley, and C. Stachniss, ``PIN-SLAM: LiDAR SLAM Using a Point-Based Implicit Neural Representation for Achieving Global Map Consistency," IEEE Trans. Robot., vol 40, pp. 4045-4064, 2024.
\bibitem{wei19} X. Wei, I. A. Bârsan, S. Wang, J. Martinez, and R. Urtasun, ``Learning to Localize Through Compressed Binary Maps," Proc. IEEE Conf. Comput. Vis. Pattern Recognit., pp. 10308-10316, 2019.
\bibitem{yin21} H. Yin, Y. Wang, L. Tang, X. Ding, S. Huang, and R. Xiong, ``3D LiDAR Map Compression for Efficient Localization on Resource Constrained Vehicles," IEEE Trans. Intell. Transp. Syst., vol. 22, no. 2, pp. 837-852, 2021.
\bibitem{chen21} X. Chen, I. Vizzo, T. Läbe, J. Behley, and C. Stachniss, ``Range Image-based LiDAR Localization for Autonomous Vehicles," Proc. IEEE Int. Conf. Robot. Autom., pp. 5802-5808, 2021.
\bibitem{yuan22} C. Yuan, W. Xu, X. Liu, X. Hong, and F. Zhang, ``Efficient and Probabilistic Adaptive Voxel Mapping for Accurate Online LiDAR Odometry," IEEE Robot. Autom. Lett., vol. 7, no. 3, pp. 8518-8525, 2022.
\bibitem{weng18} L. Weng, M. Yang, L. Guo, B. Wang, and C. Wang, ``Pole-Based Real-Time Localization for Autonomous Driving in Congested Urban Scenarios," Proc. IEEE Int. Conf. Real-Time Comput. Robot., pp. 96-101, 2018.
\bibitem{cao20} B. Cao, C. -N. Ritter, D. Göhring, and R. Rojas, ``Accurate Localization of Autonomous Vehicles Based on Pattern Matching and Graph-Based Optimization in Urban Environments," Proc. IEEE Int. Conf. Intell. Transp. Syst., pp. 1-6, 2020.
\bibitem{steinke21} N. Steinke, C. -N. Ritter, D. Goehring, and R. Rojas, ``Robust LiDAR Feature Localization for Autonomous Vehicles Using Geometric Fingerprinting on Open Datasets," IEEE Robot. Autom. Lett., vol. 6, no. 2, pp. 2761-2767, 2021.
\bibitem{shi23} P. Shi, J. Li, and Y. Zhang, ``LiDAR localization at 100 FPS: A map-aided and template descriptor-based global method," Int. J. Appl. Earth Obs. Geoinf., vol. 120, 103336, 2023
\bibitem{wolcott17} R. Wolcott and R. Eustice, ``Robust LIDAR localization using multiresolution Gaussian mixture maps for autonomous driving," Int. J. Rob. Res., vol. 36, no. 3, pp. 292-319, 2017.
\bibitem{luo22} L. Luo, S. -Y. Cao, Z. Sheng, and H. -L. Shen, ``LiDAR-Based Global Localization Using Histogram of Orientations of Principal Normals," IEEE Trans. Intell. Veh., vol. 7, no. 3, pp. 771-782, Sept. 2022. 
\bibitem{wan18} G. Wan, X. Yang, R. Cai, H. Li, Y. Zhou, H. Wang, and S. Song, ``Robust and precise vehicle localization based on multi-sensor fusion in diverse city scenes," Proc. IEEE Int. Conf. Robot. Autom., pp. 4670-4677, 2018.
\bibitem{barsan18} I. A. Bârsan, S. Wang, A. Pokrovsky, and R. Urtasun, ``Learning to Localize Using a LiDAR Intensity Map," Proc. Conf. Robot Learn., vol. 87, pp. 605-616, 2018. 
\bibitem{lu19} W. Lu, Y. Zhou, G. Wan, S. Hou, and S. Song. ``L3-net: Towards learning based lidar localization for autonomous driving," Proc. IEEE Conf. Comput. Vis. Pattern Recognit., pp. 6389-6398, 2019.
\bibitem{rozenberszki20} D. Rozenberszki, and A. Majdik, ``{LOL}: Lidar-only Odometry and Localization in 3D point cloud maps," Proc. IEEE Int. Conf. Robot. Autom., pp. 4379-4385, 2020.
\bibitem{dube19} R. Dubé, A. Cramariuc, D. Dugas, H. Sommer, M. Dymczyk, J. Nieto, R. Siegwart, and C. Cadena, ``{SegMap}: Segment-based mapping and localization using data-driven descriptors," Int. J. Robot. Res., 2019.
\bibitem{adurthi23} N. Adurthi, ``Scan Matching-Based Particle Filter for LIDAR-Only Localization," Sensors 23, no. 8:4010, 2023.
\bibitem{steinke24} N. Steinke, D. Goehring and R. Rojas, ``GroundGrid: LiDAR Point Cloud Ground Segmentation and Terrain Estimation," IEEE Robot. Autom. Lett., vol. 9, no. 1, pp. 420-426, 2024.
\bibitem{revaud19} J. Revaud, C. De Souza, M. Humenberger, and P. Weinzaepfel, ``R2D2: Reliable and Repeatable Detector and Descriptor," Adv. Neural Inf. Process. Syst., vol. 32, 2019.
\bibitem{lim22} H. Lim, S. Yeon, S. Ryu, Y. Lee, Y. Kim, J. Yun, E. Jung, D. Lee, and H. Myung, ``A Single Correspondence Is Enough: Robust Global Registration to Avoid Degeneracy in Urban Environments," Proc. IEEE Int. Conf. Robot. Autom., pp. 8010-8017, 2022.
\bibitem{fischler1981} M. A. Fischler and R. C. Bolles, ``Random sample consensus: a paradigm for model fitting with applications to image analysis and automated cartography," Commun. ACM, vol. 24, no. 6, pp. 381-195, 1981.
\bibitem{liao23} Y. Liao, J. Xie, and A. Geiger, ``KITTI-360: A Novel Dataset and Benchmarks for Urban Scene Understanding in 2D and 3D," IEEE Trans. on Pattern Anal. and Mach. Intell., vol. 45, no. 3, pp. 3292-3310, 2023.
\bibitem{kim20} G. Kim, Y. S. Park, Y. Cho, J. Jeong, and A. Kim, ``MulRan: Multimodal Range Dataset for Urban Place Recognition," Proc. IEEE Int. Conf. Robot. Autom., pp. 6246-6253, 2020.
\bibitem{geiger12} A. Geiger, P. Lenz, and R. Urtasun, ``Are we ready for Autonomous Driving? The KITTI Vision Benchmark Suite," Proc. IEEE Conf. Comput. Vis. Pattern Recognit., pp. 3354--3361, 2012.
\bibitem{guadagnio25} T. Guadagnino, B. Mersch, S. Gupta, I. Vizzo, G. Grisetti, and C. Stachniss, ``KISS-SLAM: A Simple, Robust, and Accurate 3D LiDAR SLAM System With Enhanced Generalization Capabilities," arXiv preprint, arXiv:2503.12660, 2025.
\bibitem{chen91} Y. Chen and G. Medioni, ``Object modeling by registration of multiple range images," Proc. IEEE IEEE Int. Conf. Robot. Autom., pp. 2724-2729, 1991.
\end{thebibliography}
\end{document}